\documentclass[
]{ceurart}

\usepackage{listings}
\usepackage{url}

\begin{document}

%%
%% Rights management information.
%% CC-BY is default license.
\copyrightyear{2026}
\copyrightclause{Copyright for this paper by its authors.
  Use permitted under Creative Commons License Attribution 4.0
  International (CC BY 4.0).}

%%
%% This command is for the conference information
\conference{EVALITA 2026: 9th Evaluation Campaign of Natural Language
Processing and Speech Tools for Italian, Feb 26 – 27, Bari, IT}

%%
%% The "title" command
\title{Peacemaker at ATE-IT: Automatic term extraction from Italian text for waste management data using  encoder model}

%%
%% The "author" command and its associated commands are used to define
%% the authors and their affiliations.
\author[1]{Mahdi Bakhtiyarzadeh}[%
orcid=,
email=m.bakhtiyarzadeh1403@ms.tabrizu.ac.ir,
url=https://github.com/Mahdi8424,
]
\cormark[1]
%\fnmark[1]
\address[1]{Department of Computer Science, University of Tabriz,
29 Bahman Boulevard,
Tabriz 51666-16471, Iran}

\author[2]{Hadi Bayrami Asl Tekanlou}[%
orcid=0009-0002-7206-6116,
email=h.bayrami1403@ms.tabrizu.ac.ir,
url=https://github.com/HadiBayrami/,
]
\cormark[1]
%\fnmark[1]
\address[2]{Department of Computer Science, University of Tabriz,
29 Bahman Boulevard,
Tabriz 51666-16471, Iran}
\author[1]{Jafar Razmara}[%
orcid=0000-0002-6320-8517,
email=razmara@tabrizu.ac.ir,
url=,
]
\cormark[1]
%\fnmark[1]
\address[]{}

%% Footnotes
\cortext[1]{Corresponding author.}
\fntext[1]{These authors contributed equally.}

%%
%% The abstract is a short summary of the work to be presented in the
%% article.
\begin{abstract}
% Automatic term extraction plays a pivotal role in today’s technology-focused world, as it is widely %used in many modern search engines. While recent advances show promising performance, accurately %labeling domain-specific terms remains challenging due to multiple factors, ranging from the lack of %annotated data to complex multi-word expressions and domain shift. In this work, we introduce a %lightweight and interpretable approach for automatic term extraction, developed for Task A of the ATE %Shared Task. The proposed method relies on fine-tuning–based extraction strategies and is designed to %operate under limited computational resources. We assess our system using both type-level and micro-%level precision, recall, and F1-score to reflect complementary aspects of extraction performance.
The development of automatic term extraction has become increasingly important in modern technology. Automatic term extraction can be found in virtually every search engine that is currently available to users. Recent advancements have provided promising results for the extraction of automatic terms; however, accurate labeling is difficult because of several factors, such as the limited number of annotated documents available for training and the complexity of extracting multi-word expressions due to shifts in the domain. In this paper, we will present a low-cost and interpretable method of automatic term extraction, developed specifically for Task A of the ATE Shared Task. This new method utilizes fine-tuning extraction strategies that can run on a small amount of computational resources. We evaluated our automated system using both type-level and micro-level measures of precision, recall, and F1-score to measure both complementary aspects of the extraction performance. According to the experimental results, our proposed approach achieves consistent and balanced performance compared to other teams. Even though the technique itself is relatively straightforward, it serves as a good starting point for low-resource models. Overall, the findings point toward the possibility of significant future advancements (in model expansion) with higher-level performance still able to retain their ability to be interpreted.

\end{abstract}

%%
%% Keywords. The author(s) should pick words that accurately describe
%% the work being presented. Separate the keywords with commas.

%%
%% This command processes the author and affiliation and title
%% information and builds the first part of the formatted document.

\maketitle

\section{Introduction}

Terminology is the beating heart of sentences, especially in tasks such as Automatic Term Extraction (ATE), where identifying domain-specific concepts from corpora is crucial for applications like machine translation, knowledge extraction, and semantic analysis~\cite{10.1007/978-3-031-70239-6_10,10.1093/llc/fqad030}. The "Automatic Term Extraction" (ATE) focuses on tasks such as corpus construction, unithood (measuring word co-occurrence), termhood (assessing domain relevance), and recognizing "variants" or how a term can be expressed differently from the original way~\cite{Heylen2014AutomaticTE}. Recent research utilizing large language models for few-shot ATE demonstrated considerable progress when operated under conditions where limited resources exist. Additionally~\cite{10.1007/978-3-031-70239-6_10}, systematic reviews have documented ongoing development of methods for extracting multi-word terms from specific types of text through Transformer's architecture and ability to extract terms across multiple languages and domains~\cite{10.1093/llc/fqad030,lang-etal-2021-transforming}. These advances play a significant role in the field of computer-assisted translation (CAT), where having bilingual terms available improves the quality of machine translations~\cite{arcan-etal-2014-enhancing} as well as supporting the evaluation of term extraction software against gold standard datasets such as ACLRD-TEC2.0 and MAGMATic~\cite{qasemizadeh-schumann-2016-acl,scansani-etal-2019-magmatic}. Monolingual and Multilingual ATE~\cite{RigoutsTerryn2020}, extending into specialized fields such as biomedical literature~\cite{Tiwari2020-my} and media bias analysis~\cite{spinde2025leveraginglargelanguagemodels}, is supported by datasets created from comparable corpora. It is common practice when improving ATE to integrate terminological resources~\cite{aubin2006improvingtermextractionterminological} with well defined robust evaluation frameworks~\cite{article}. However, challenges still exist regarding the collection of semi-automated data for Systematic Reviews~\cite{Schmidt2021-lq}.

This study makes the following contributions:

\begin{itemize}
    \item \textbf{Robust Data Alignment and Preparation}: We implement a custom preprocessing pipeline featuring token-level semantic alignment (BIO\-tagging), specifically tailored for the Italian language and the ATE-IT task. This process accurately maps and converts multi\-word term instances from the raw gold standard into the correct B\-TERM, I\-TERM, O sequence labels, resolving token boundary ambiguities essential for effective Token Classification.
    \item \textbf{High-Fidelity Feature Extraction Architecture}: We utilize a high-performance, context-aware Transformer-based Token Classification architecture, leveraging the pre-trained dbmdz/bert-base-italian-cased model. This provides deep, contextual semantic representations optimized for token-level predictions within the Italian regulatory domain.
    \item \textbf{Environment Stability and Reproducibility}: We ensure the stability and reproducibility of the fine-tuning process by explicitly pinning critical Python dependencies (including transformers and accelerate) to verified, compatible versions. This guarantees API compatibility across varying runtime environments, mitigating common failure modes and allowing for reliable training execution.

\end{itemize}

\section{Related Work}

As presented in the previous section, Automatic Term Extraction (ATE) is critical for extracting domain-specific concepts that underpin the core tasks of natural language processing and knowledge management. The initial strategies that dealt with ATE were based on the combination of linguistic and statistical techniques to address issues such as unithood (the occurrence of words together) and termhood (the relevance of terms to domains). Examples of hybrid methods used on specialized corpora, such as in computer science and medicine, include the combination of statistical measures (mutual information and C-value/NC-value) with shallow linguistic criteria (part of speech)~\cite{Milios2003AUTOMATICTE}. Various approaches have shown strong performance across domains, with similarity scores closely matching those for terms represented via traditional word-based representations in document similarity evaluations. However, the new methods will need to compensate for cases where multi-word terms have nesting problems~\cite{Milios2003AUTOMATICTE}.  Additionally, the comparative evaluations of term recognition systems indicate that combining several different approaches, such as using voting methods, resulted in better performance than using any one method alone when evaluated against corpus data from Wikipedia text, but performance varied from domain to domain and was not as high when using data from the Genia corpus in the life sciences~\cite{zhang-etal-2008-comparative}. Tools based on shallow grammars designed for use with noun-phrases in technical circles and combined with the application of clustering to accommodate variations in their structure (e.g., using variation of insertion structure) have generated very high levels of precision ranging from 93\% to 98\% when constructing an automated thesaurus~\cite{10.3115/977035.977039}. PAT-tree approach also has been shown to enable the construction of adaptive keyword extraction systems, reducing dependency on rigid lexicons and word segmentation of documents written in Asian languages, such as Chinese, and facilitating their effective retrieval or classification when using these systems~\cite{10.1145/278459.258534}. Multilingual and Monolingual ATE advancements take advantage of comparable corpora to develop data sets that can be used to support variant recognition and cross-lingual transfer. One particular contribution is the development of a data set specifically for both monolingual and multilingual ATE that allows terms to be extracted from comparable sources and resolve resource scarcity in low-resource languages~\cite{RigoutsTerryn2020,10.3115/977035.977039}. This reflects an interest in leveraging corpus-based semantic connections to create associations to MWEs by way of Information Extraction, and using these associations to compare the MWEs to thesauri within the Agriculture and Medicine domains~\cite{10.3115/1034678.1034739}. For Agriculture, rule-based systems like RelExOnt have automated vocabulary extraction as well as relation discovery, providing ontologies with an 86.89\% accuracy rate in the creation process~\cite{KAUSHIK201860}. The combination of statistical, linguistic, and hybrid extractors has increased the accuracy of Spanish medical text extraction. It demonstrates the value of ensemble methods for different extraction methods~\cite{article}. Recent advancements in medical text extraction systems incorporating complete terminological resources and comprehensive evaluation methods allow for greater reliability of ATE systems (Automated Term Extraction). Incorporating external lexicons specific to the domain and other validated sources of information into automation models for extraction provide significant increases in recognizing multifaceted terms~\cite{10.3115/1034678.1034739}.

\section{Task Description and Dataset}
\subsection{Task}
This section describes the parameters for the ATE-IT Shared Task 2026 and its components, as well as the dataset used in the evaluation. It also describes the system architecture created for use in the evaluation; our contribution was in \textbf {Subtask A: Term Extraction}.
\subsubsection{ATE-IT Shared Task 2026: Subtasks and Primary Focus}
The ATE-IT Shared Task 2026~\cite{cirillo2026ateit} has been organized around an evaluation of Automated Term Extraction (ATE) in a large scale way at the specific domain level of waste management in Italy.  The challenge is divided into two distinct subtasks of increasing complexity:

\begin{enumerate}
    \item \textbf{Subtask A: Term Extraction}. The main goal in the task of Term Extraction is to identify and extract individual domain-specific terms from the corpus sentences.

    \item \textbf{Subtask B: Term Variants Clustering.} The primary objective of Term Variants Clustering is to group extracted terms that refer to the same underlying concept. This requires semantic and morphological analysis of terms so that synonymy and lexical variations can be handled.
    
\end{enumerate}
In this study, we will be participating only in Subtask A: Term Extraction. Our main goal will be to process the sentences of the specialized corpus accurately and identify the municipal waste management domain-specific terms.

\subsection{Corpus Description}

Ate-it challenge organizers supplied the official textual source for the ATE-IT shared task, which contained textual examples related directly to the field of municipal waste management (an area of environmental concern). The challenge also provided guidelines regarding how to use the corpus dataset for this challenge, in order to develop models based on the provided training data and evaluate models against a pre-defined testing dataset in accordance with the established structure of the challenge. Tables~\ref{stats} and~\ref{rep} present an overview of the ATE-IT data statistics and representative examples from the dataset, respectively.

\begin{table}[h]
\caption{\label{stats} ATE-IT data statistic overview.}
\begin{center}
\begin{tabular}{|l|l|}
\hline 
\bf Dataset & \bf Samples  \\ 
\hline
Train & 2308 \\
\hline
Development & 577 \\
\hline
Test & 1142\\

\hline
\end{tabular}
\end{center}

\end{table}

\begin{table}[h]
\centering
\caption{Representative examples from the dataset}
\label{rep}

\begin{tabular}{p{10cm} p{5cm}}
\hline
\textbf{sentence\_text} & \textbf{term\_list} \\
\hline
CONSIDERATO che la situazione in cui si trova attualmente il Comune
è riconducibile all'ipotesi contemplata nelle previsioni di cui al citato
art. 191, in quanto sussistono gravi condizioni e fondate ragioni di tutela
della salute pubblica e dell'ambiente, che risulterebbero inevitabilmente
pregiudicate in caso di mancato ricorso temporaneo ad una speciale forma
di gestione del servizio di raccolta per i rifiuti provenienti da luoghi
adibiti ad uso di civile abitazione in cui alloggino persone risultate
positive al Covid-19, in quarantena obbligatoria ai sensi dell'articolo 1
lettera e del D.P.C.M. 8 marzo 2020
&
``gestione del servizio di raccolta per i rifiuti'',
``servizio di raccolta per i rifiuti provenienti da luoghi adibiti ad uso
di civile abitazione''
\\
\hline
Al fine di consentire la raccolta dei rifiuti nei contenitori interni agli
stabili, il proprietario singolo o i condomini, in solido fra loro, hanno
l'obbligo di esporre gli stessi nei giorni e nelle ore stabiliti sul tratto
viario prospiciente l'immobile di competenza e di riporli all'interno dei
cortili o delle pertinenze condominiali, dopo l'avvenuto servizio di raccolta.
&
``esporre'',
``raccolta dei rifiuti'',
``servizio di raccolta''
\\
\hline
La richiesta di attivazione del servizio deve essere presentata dall'utente al gestore dell'attivit\u00e0 di gestione tariffe e rapporto con gli utenti entro novanta (90) giorni solari dalla data di inizio del possesso o della detenzione dell'immobile, a mezzo posta, via e-mail o - 5 mediante sportello fisico e online di cui all'Articolo 19, compilando 2 l'apposito modulo scaricabile dalla home page del sito internet del 01 gestore in modalit\u00e0 anche stampabile, disponibile presso gli sportelli fisici, laddove presenti, ovvero compilabile online art. &                 "gestore", "tariffe", "utente"\\
\hline

\end{tabular}
\end{table}

\section{Proposed Approach}
The system that has been developed to complete ATE\_IT Subtask A (extraction of Italian vocabulary) uses a Sequence Labeling Framework. The identification of terms has been changed into classifying every single word in a sentence. We used a fine-tuned state-of-the-art transformer model for the extraction of Vocabulary.

\subsection{Data Preprocessing and Alignment}
A critical component of the methodology is the custom preprocessing pipeline, which converts raw, sentence-level term annotations into a format consumable by the sequence labeling model.
\begin{itemize}
    \item \textbf{BIO Tagging Scheme}: The standard B-I-O (Begin, Inside, Outside) tagging scheme is employed: \texttt{B-TERM} for the first token of a term, \texttt{I-TERM} for subsequent tokens, and \texttt{O} for non-term tokens.

    \item \textbf{Token-Term Alignment}: Term-to-token correspondence is ensured by using offset mappings provided by the tokenizer to create a mapping between character-level terms and the subword tokens from the model. To avoid double labels and overlap between terms that are themselves "overlapping," a mask is created for each character in the input sentence as it emerges from the model for each term. The mask makes sure that for every character position, it can only be mapped to one term, thus ensuring that a single character can't be allocated to two separate terms.

    \item \textbf{Label Encoding}: Symbolic BIO labels are converted into numerical IDs (O=0, B-TERM=1, I-TERM=2). Special tokens (e.g., [CLS], [SEP]) and padded tokens are masked with a label ID of \textbf{-100}, which is ignored by the loss function.

\end{itemize}
\subsection{Modeling and Training}
\begin{figure*}[t]
   \centering
   \includegraphics[width=0.8\textwidth]{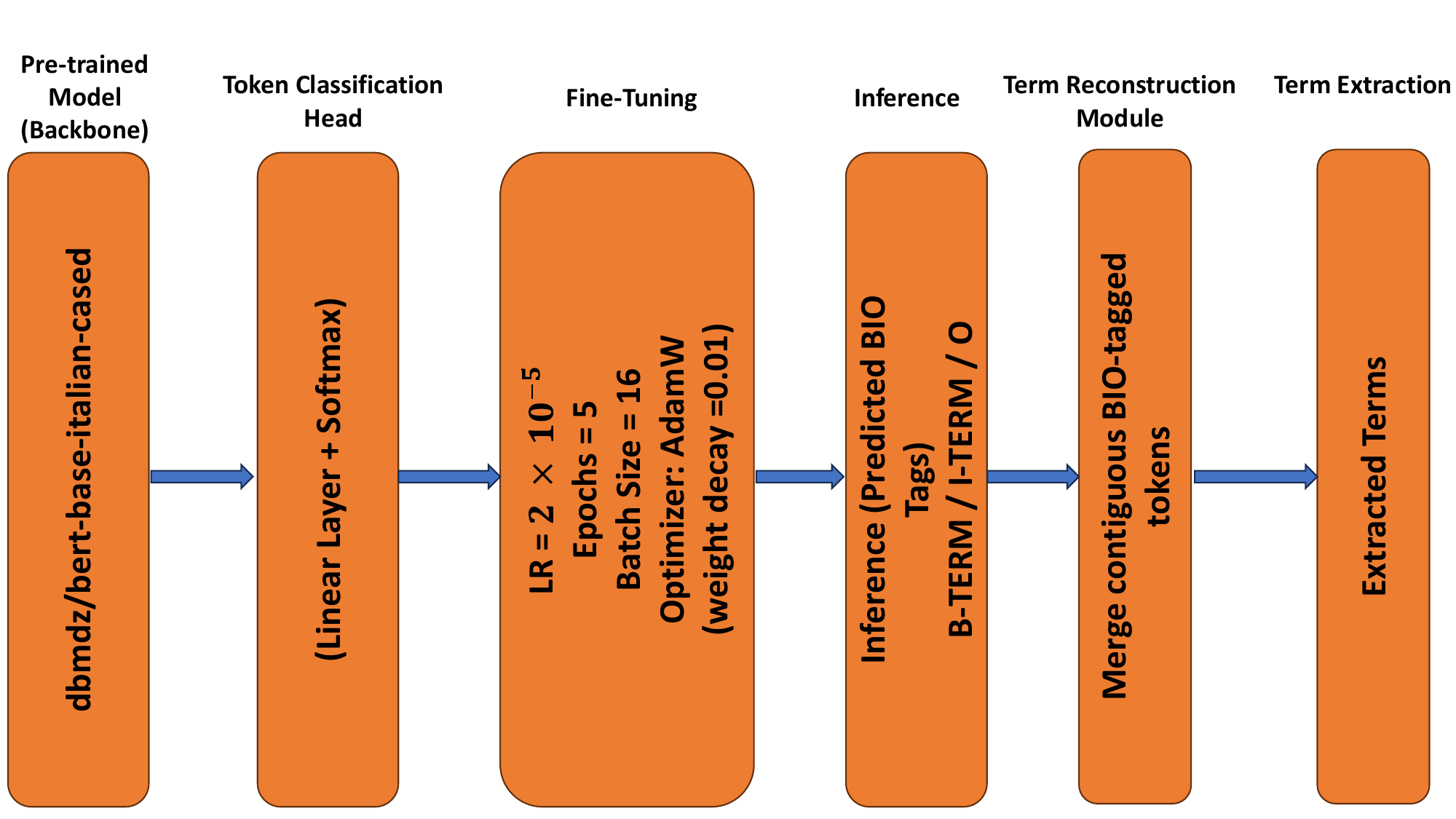}  
   \centering \caption{Overview of the system architecture used for token-level term extraction.}
\label{fig:architecture}
\end{figure*}

The core of the system is the \textit{dbmdz/bert-base-italian-cased} model, a cased \textbf{BERT} (\textbf{Bidirectional Encoder Representations from Transformers}) variant pre-trained on an extensive Italian corpus. This model is adapted for \textbf{Token Classification} by appending a linear layer for transfer learning.\\
The system architecture is illustrated in Figure~\ref{fig:architecture}.

%% The declaration on generative AI comes in effect
%% in Janary 2025. See also
%% https://ceur-ws.org/GenAI/Policy.html

\section{Results}
In this section, we describe the results obtained in the Automatic Term Extraction 2026 shared task. A total of 9 teams participated in the competition. The present paper presents results based on micro and type based evaluation metrics. Micro measures consider the total count of each type of term (e.g., phone numbers, names) that appear in the data set and provide a measure of accuracy of the entire extraction process. Type measures provide insight into how well the system is able to find different types of terms within the dataset.

\subsection{Baselines}

As a reference point, Table~\ref{baseline_results} presents the baseline results for both type-based and micro-based evaluation metrics. The following is our first baseline, which should be used to compare all participating systems. This baseline was created using the latest version (gemini-2.5-flash) of a closed-source large language model, in a zero-shot mode. The input text was processed in 20 instance batches according to the predefined prompt.

\begin{table}[h]
\caption{\label{baseline_results} Baseline model results using type-based and micro-based evaluation metrics}
\begin{center}
\begin{tabular}{|l|l|l|l|l|l|}
\hline 
\bf Type Precision & \bf Type Recall & \bf Type F1 & \bf Micro Precision & \bf Micro Recall & \bf Micro F1 \\ 
\hline
0.435 & 0.508 & 0.469 & 0.497 & 0.559 & 0.526 \\
\hline
\end{tabular}
\end{center}
\end{table}

\subsection{Evaluation Metrics}
\label{sec:metrics}
The criteria used to evaluate the submitted work included a comparison between the automated term extraction and the corresponding gold standard annotations (as previously outlined). At the same time, type-level and micro-level metrics provide separate initial views of system performance. Type-level metrics provide an aggregate measure of how many unique domain terms have been identified by the system regardless of their frequency of occurrence; micro-level metrics provide detailed performance data on a sentence-by-sentence basis by accumulating all "True," "False Positive," and "False negative" results from all sentences to arrive at totals. Additionally, the precision, recall and F1 score calculations were derived from both type-level and micro-level evaluations allowing for assessment between extracted term correctness and completeness from both perspectives.\\
\\
\bf Type Metrics
%\begin{itemize}
  %\item Precision
  \begin{itemize}
      \item TP = number of extracted unique terms that appear in the gold set
      \item FP = number of extracted unique terms that do not appear in the gold set
      \item FN = number of gold unique terms that were not extracteds
  \end{itemize}
  \[
    Precision = \frac{TP}{TP + FP}
  \]

  %\item Recall
  \[
    Recall = \frac{TP}{TP + FN}
  \]

 % \item F1
  \[
    F1 = \frac{2 \cdot Precision \cdot Recall}{Precision + Recall}
  \]
%\end{itemize}

\bf Micro Metrics~\cite{doi:10.3233/SW-170286}

\begin{itemize}
  \item $\boldsymbol{\mathrm{TP}}_s$ = number of terms extracted from sentence $s$ that match the gold standard
  \item $\boldsymbol{\mathrm{FP}}_s$ = number of terms extracted from sentence $s$ that do not match the gold standard
  \item $\boldsymbol{\mathrm{FN}}_s$ = number of gold standard terms in sentence $s$ that were not extracted
\end{itemize}

\[
Precision_{\text{micro}} =
\frac{\sum_{s \in D} TP_s}
     {\sum_{s \in D} (TP_s + FP_s)}
\]

\[
Recall_{\text{micro}} =
\frac{\sum_{s \in D} TP_s}
     {\sum_{s \in D} (TP_s + FN_s)}
\]

\[
F1_{\text{micro}} =
\frac{2 \cdot Precision_{\text{micro}} \cdot Recall_{\text{micro}}}
     {Precision_{\text{micro}} + Recall_{\text{micro}}}
\]

\normalfont\normalsize

\subsection{Team Results}

\begin{figure*}[b]
  \centering
  \label{radar}
  \includegraphics[width=0.8\textwidth]{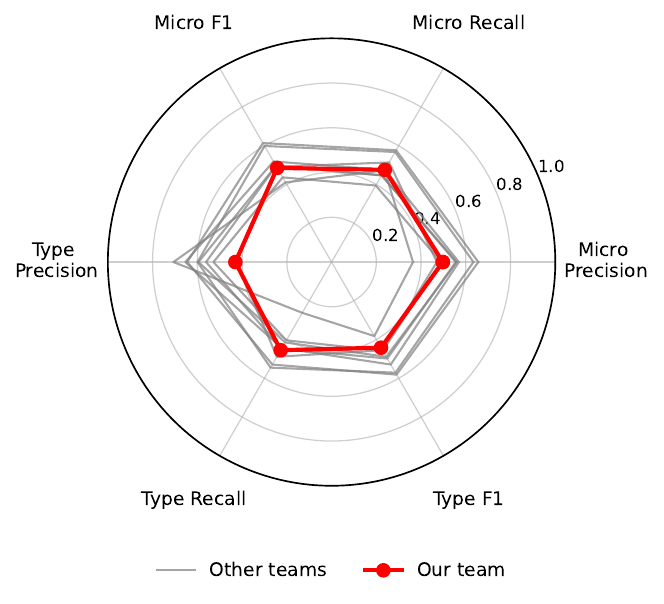}
  \caption{Radar plot comparing the performance of our approach with other participating systems across type-level and micro-level Precision, Recall, and F1-score.}
  \label{fig:radar}
\end{figure*}

The outcomes from our application of the method we’ve described will now be discussed in detail. The performance of our system is assessed with the metrics shown in Section~\ref{sec:metrics}, and as for how our performance measures up against the results reported by each of the other teams that participated in this study, it is important to first compare these results so that we can accurately identify where the submitted systems rank with respect to one another and any associated strengths/weaknesses. For reference, Table~\ref{peacemaker_results} presents the detailed type-level and micro-level evaluation metrics achieved by the Peacemaker team, which are used as a baseline comparison in our analysis. In order to compare our model's performance against all evaluated dimensions, we provide a multi-dimensional view of our model's performance along with those of its competitors through the use of a radar plot (Figure~\ref{fig:radar}), which allows us to quickly compare Precision, Recall and F1-score evaluations for the two evaluation types (type-level and micro-level) evaluated.

\begin{table}[h]
\caption{\label{peacemaker_results} Peacemaker team results using type-based and micro-based evaluation metrics}
\begin{center}
\begin{tabular}{|l|l|l|l|l|l|}
\hline 
\bf Type Precision & \bf Type Recall & \bf Type F1 & \bf Micro Precision & \bf Micro Recall & \bf Micro F1 \\ 
\hline
0.430 & 0.455 & 0.442 & 0.497 & 0.476 & 0.486 \\
\hline
\end{tabular}
\end{center}
\end{table}

\section{Discussion}

The results of the proposed method provide valid insight into how lightweight and low-engineering techniques behave in the domain of ontology term extraction even though it is not producing highest performance level. Evidence to support this claim is captured in the radar plot, which displays balanced precision and recall sections between the type-level and micro-level metrics. Therefore, this suggests that the proposed method produces stable extraction as opposed to maximally optimizing precision scores through repetitive extraction cycles.

While our experimental results show that the proposed method ranks lower than others, we have observed that for both the Type and Micro levels the Precision and Recall values are similar, indicating that this method does not rely heavily on just using high-frequency terms. We conclude that our proposed method has consistent extraction performance for both frequent and less frequent terms from many different evaluation granularities. Thus, our proposed method provides a strong baseline for future researchers in this area with a reasonable level of computational and implementation complexity.

It should be noted that our system was created using limited resources for both computation and modeling, which prevented us from being able to take full advantage of many of the larger or more specialized models that are available to some teams that have a higher ranking. If the teams were able to use better models and were able to increase their computational capabilities, it is likely that their overall performance—especially Recall—would also improve greatly. However, the current results show that the proposed method is effective and competitive and has the advantage of being lightweight. 
\section{Conclusions}

In this research, we proposed an easy-to-understand, lightweight process for extracting ontology terms using data captured during the shared task evaluation phase and applying type-level and micro-level metrics to evaluate the performance of our solution. Our reduced resources and choice of more constrained models did not hinder our application from demonstrating steady and well-balanced extraction results across all models and evaluations, which implies that the extraction capabilities of our modeling process are not based solely on selecting terms that appear very frequently.
The fact that our team came in 7th place out of the 9 teams that competed shows how well we stack up against other teams, even though we had far fewer resources than most of the other teams that took part in the competition; therefore, our system remains highly competitive based on what we have demonstrated.

\section*{Declaration on Generative AI}
 Generative AI tools played a supplemental and shallow role in our work. We used generative AI for brainstorming and developing ideas on how to tackle the Automatic Term Extraction (ATE) task; they were also helpful for enhancing the clarity and organization of notes and drafts. However, we did not rely on generative AI to create any part of the dataset, including the annotations, experimental results, or final models. All authors made the decisions about method choice, implementation, and analysis. After using these tools/services, the authors reviewed and edited the content as needed and takes full responsibility for the publication’s content.

\bibliography{refs}
\end{document}